\documentclass[10pt,twocolumn,letterpaper]{article}

\usepackage{iccv}
\usepackage{times}
\usepackage{epsfig}
\usepackage{graphicx}
\usepackage{amsmath}
\usepackage{amssymb}
\usepackage{multicol}
\usepackage{graphicx}
\usepackage{amssymb}
\usepackage{pifont}
\usepackage{booktabs}
\usepackage{mwe}
\usepackage{array}
\usepackage[dvipsnames]{xcolor}
\usepackage{courier}
\usepackage{float}
\usepackage{dblfloatfix}
\usepackage[accsupp]{axessibility}

\newcommand{\cmark}{\ding{51}}%
\newcommand{\xmark}{\ding{55}}%
\newcolumntype{M}[1]{>{\centering\arraybackslash}m{#1}}

\usepackage[pagebackref=true,breaklinks=true,letterpaper=true,colorlinks,bookmarks=false]{hyperref}

\iccvfinalcopy 

\def\iccvPaperID{1408} 

\ificcvfinal\fi

\begin{document}

\title{Human Preference Score: \\ Better Aligning Text-to-Image Models with Human Preference}

\author{
    Xiaoshi Wu$^{1}$, 
    Keqiang Sun$^{1}$, 
    Feng Zhu$^{2}$, 
    Rui Zhao$^{2, 3}$, 
    Hongsheng Li$^{1, 4, 5}$
\vspace{0.1em}\\
    $^{1}$Multimedia Laboratory, The Chinese University of Hong Kong \\
    $^{2}$SenseTime Research \quad $^{3}$Qing Yuan Research Institute, Shanghai Jiao Tong University \\
    $^{4}$Centre for Perceptual and Interactive Intelligence (CPII) \quad
    $^{5}$Shanghai AI Laboratory \\
    \texttt{\small \{wuxiaoshi@link, kqsun@link, hsli@ee\}.cuhk.edu.hk},~\texttt{\small \{zhufeng, zhaorui\}@sensetime.com}
}

\maketitle
\ificcvfinal\thispagestyle{empty}\fi

\begin{abstract}
Recent years have witnessed a rapid growth of deep generative models, with text-to-image models gaining significant attention from the public. 
However, existing models often generate images that do not align well with human preferences, such as awkward combinations of limbs and facial expressions. 
To address this issue, we collect a dataset of human choices on generated images from the Stable Foundation Discord channel. 
Our experiments demonstrate that current evaluation metrics for generative models do not correlate well with human choices.
Thus, we train a human preference classifier with the collected dataset and derive a Human Preference Score (HPS) based on the classifier. 
Using HPS, we propose a simple yet effective method to adapt Stable Diffusion to better align with human preferences. 
Our experiments show that HPS outperforms CLIP in predicting human choices and has good generalization capability toward images generated from other models. 
By tuning Stable Diffusion with the guidance of HPS, the adapted model is able to generate images that are more preferred by human users.
The project page is available here: \href{https://tgxs002.github.io/align\_sd\_web/}{https://tgxs002.github.io/align\_sd\_web/}.

\end{abstract}

\section{Introduction}

\begin{figure}
    {
    \centering
    \includegraphics[width=1.\linewidth]{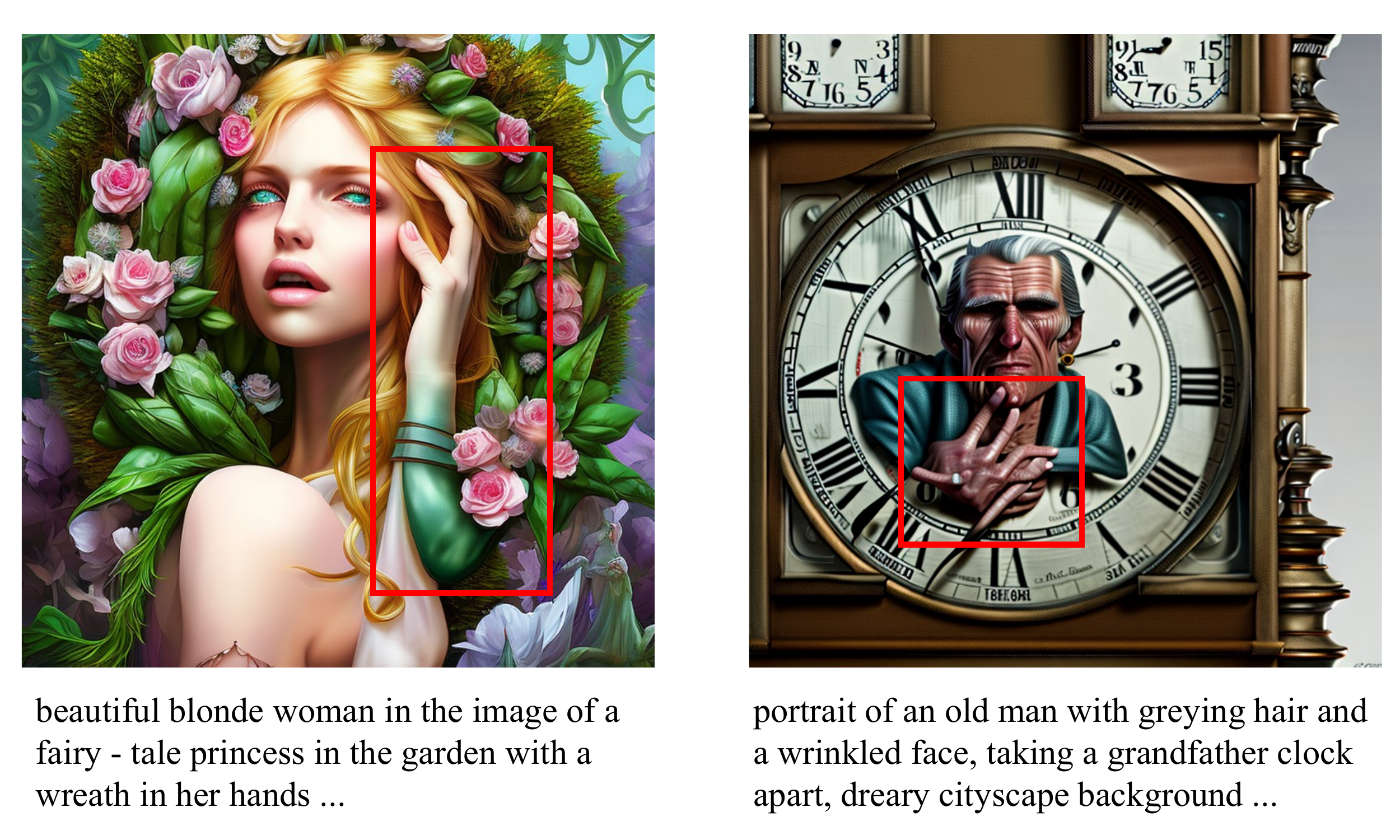} 
    \caption{Generated images often do not align well with human preferences and intentions. Input prompts are shown below images.}
    \label{fig:seed}
    }
    {
    \centering
    \includegraphics[width=1.0\linewidth]{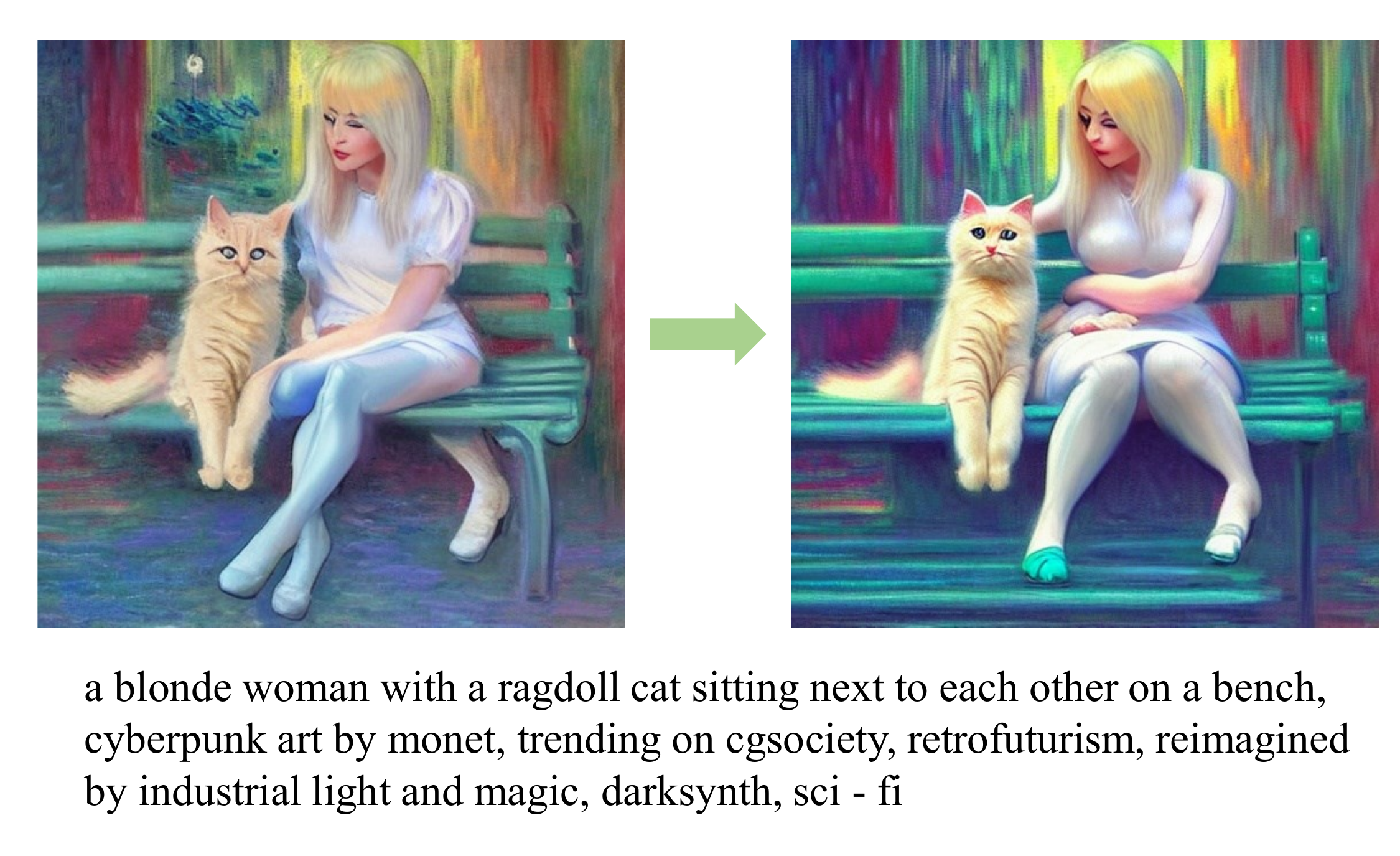}
    \caption{We show that Stable Diffusion v1.4 can be adapted to better align with human preferences and intentions when guided by the proposed human preference classifier. The input prompt is shown below images.}
    }
\end{figure}

The recent progress in diffusion models~\cite{glide, dalle2, imagen, stable_diffusion} has enabled impressive advancements in text-to-image generation, with many models now being deployed in real-world applications such as DALL·E~\cite{dalle2} and Stable Diffusion~\cite{stable_diffusion}. 
However, public attention has also highlighted new issues, such as the awkward combinations of limbs and facial expressions of generated persons as shown in Fig.~\ref{fig:seed}. 
The users usually need to cherry-pick results to avoid these artifacts.
In other words, the generated images are misaligned with human preferences.


To further improve the quality of generated images, it is essential to track the ability of a model to generate human preferable images. 
However, it is uncertain whether the existing evaluation metrics, such as Inception Score (IS)~\cite{inception_score} and Fréchet inception distance (FID)~\cite{fid}, are correlated with human choices. 
These metrics perceive an image through a classification-based CNN trained on ImageNet~\cite{imagenet}, which has been shown to be biased towards image texture rather than general image contents ~\cite{Geirhos2018ImageNettrainedCA}, and thus may not align well with human perception. 
Also, both IS and FID are single-modal evaluation metrics, which do not take user intention into account.
Some recent studies~\cite{glide, dalle2, instructpix2pix} use the CLIP~\cite{clip} model as a proxy for human judgment to evaluate the alignment between generated images and text prompts. 
The CLIP~\cite{clip} model is trained on a rich dataset and is believed to capture subtle aspects of human intention better. 
However, it is uncertain whether CLIP~\cite{clip} can measure the quality of generated synthetic images, which may not adhere to the same constraints as real images, such as the example shown in Fig.~\ref{fig:seed}.

In this study, we investigate the problem of human preference using a novel, large-scale dataset of human choices on images generated by Stable Diffusion~\cite{stable_diffusion} using the same prompt.
The dataset comprises 98,807 diverse images generated from user-provided prompts, along with 25,205 human choices. 
By evaluating on this dataset, we find that the Inception Score (IS)~\cite{inception_score}, the Fréchet Inception distance (FID)~\cite{fid} and the CLIP score does not fully match the human choice, which means that the human preference is a missing dimension of image quality that is not well tracked by existing mainstream metrics.

We further train a human preference classifier on this dataset by fine-tuning the CLIP~\cite{clip} model and define human preference score (HPS) based on it. We validate HPS's alignment with human choices and its generalization capability towards other generative models through user studies. 
HPS can be utilized to guide generative models toward producing human-preferred images. 
To this end, we devise a simple yet effective method to adapt Stable Diffusion~\cite{stable_diffusion} by LoRA~\cite{lora} with awareness of human preference.
We conduct user studies to validate the effectiveness of our approach.
The results show that the adapted model can better capture human intentions, and generate more preferable images, which significantly mitigates the kind of artifact shown in Fig.~\ref{fig:seed}.

Our contributions are as follows:
(1) We create a large-scale dataset for studying human preferences.
To our best knowledge, this dataset is the first of its kind that contains massive human choices on images generated with the same prompt.
(2) We find that human choices cannot be accurately predicted by the existing mainstream evaluation metrics, while it can be better predicted via fine-tuning CLIP on the proposed dataset.
(3) We propose a simple yet effective method to guide the Stable Diffusion model toward generating images with better aesthetic quality and better alignment with human intention.

\begin{figure*}
    \centering
    \includegraphics[width=0.75\linewidth]{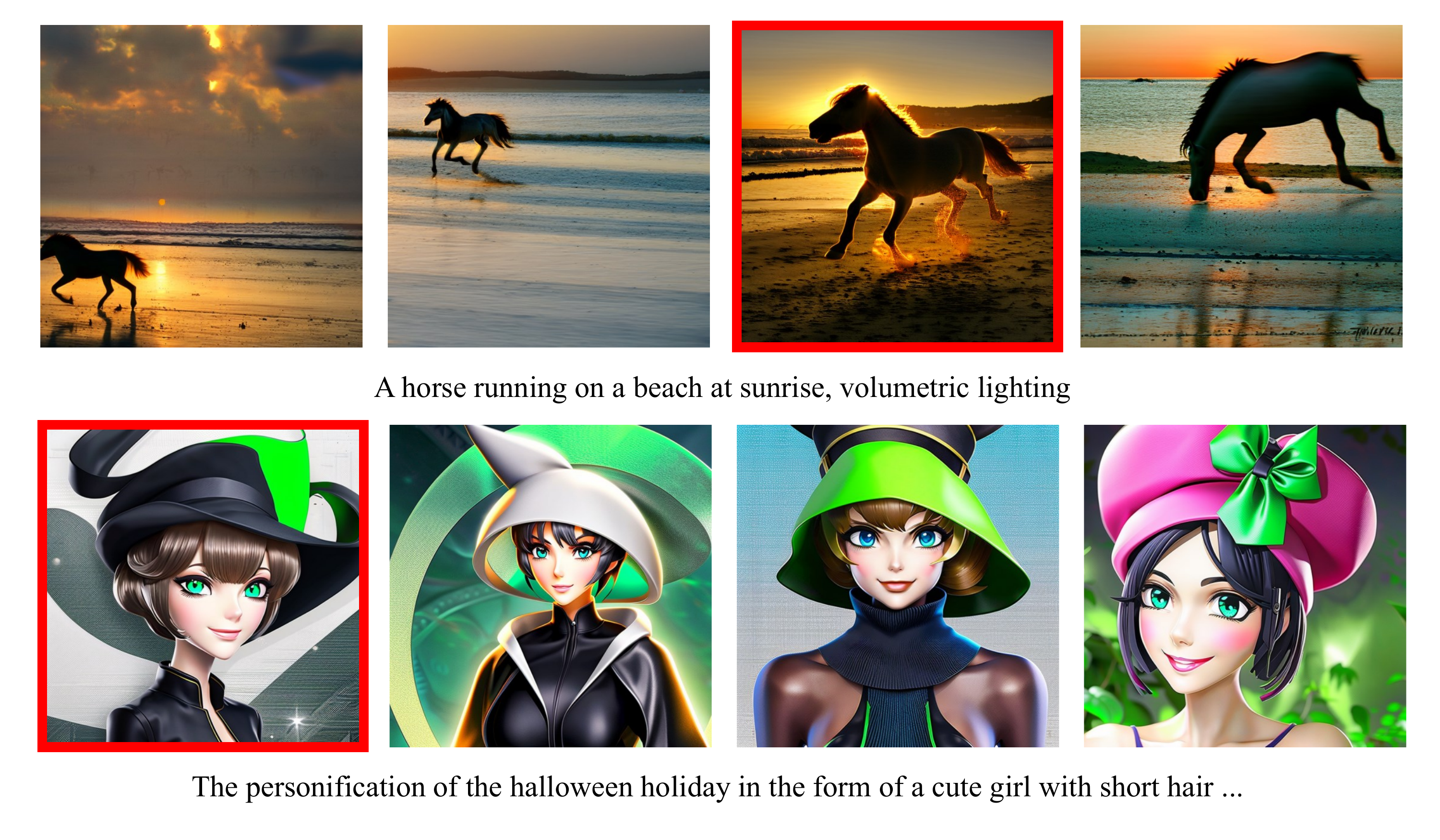}
\vspace{-0.2cm}
\caption{Examples of the collected data. The images are generated by Stable Diffusion, with corresponding prompts shown below each row of images. The preferred images are highlighted with red borders. More examples can be found in the appendix.}
\label{fig:example}
\end{figure*}

\section{Related Works}
\noindent\textbf{Text-to-image generative models.}
Text-to-image generative models have long been an active research area.
Mansimov \etal~\cite{Mansimov2015GeneratingIF} show that Deep Recurrent Attention Writer (DRAW)~\cite{Gregor2015DRAWAR} can be conditioned on captions to generate novel scene compositions.
Generative Adversarial Networks (GANs) improve image fidelity by training a discriminator to provide supervision for the generative model.
DALL·E~\cite{dalle} firstly achieves open-domain text-to-image synthesis with the help of massive image-text pairs.

Diffusion models formulate the generative process as the inverse of the diffusion process~\cite{SohlDickstein2015DeepUL}, which was improved by Song and Ermon ~\cite{Song2019GenerativeMB} and Ho \etal ~\cite{Ho2020DenoisingDP}.
Dhariwal \etal firstly show the superiority of diffusion models over GANs on image generation.
Several following works, including DALL·E 2~\cite{dalle2}, GLIDE~\cite{glide}, Imagen~\cite{imagen}, ERNIE-ViLG~\cite{feng2023ernie, Zhang2021ERNIEViLGUG}, Stable Diffusion~\cite{stable_diffusion}, bring the magic of text-to-image generation to the public attention.
Among these models, Stable Diffusion is an open-source model with an active user community.

Several recent works improve Stable Diffusion on different aspects.
DreamBooth~\cite{Ruiz2022DreamBoothFT} and ELITE~\cite{Wei2023ELITEEV} explore customizing Stable Diffusion to a certain object.
Feng \etal ~\cite{Feng2022TrainingFreeSD} propose a training-free method to guide diffusion models for better compositional capabilities.
It has been discovered that prompt engineering plays an important role in generating high-quality images.
Hao \etal ~\cite{Hao2022OptimizingPF} devise an automatic prompt engineering scheme via reinforcement learning.
Our method focuses on the misalignment between the generated image and human preference, which is orthogonal to the above-mentioned topics.

\noindent\textbf{Datasets of generated images.}
Datasets of generated images play a vital role in computer vision tasks that has difficulty in ground-truth acquisition, such as optical flow estimation~\cite{MIFDB16, DFIB15, sintel, teed2020raft, huang2022flowformer, shi2023videoflow, shi2023flowformer++}.
Thanks to active user communities of text-to-image models, several databases of images generated by diffusion models have been introduced.
Lexica (\href{https://lexica.art/}{lexica.art}) is a large database of images generated by Stable Diffusion and Lexica Aperture.
It also provides related information about the image, such as the prompt and guidance scale.
However, the database is closed-source and only allows online browsing.
DiffusionDB~\cite{diffusiondb} is a large-scale open-source database collected from the Stable Foundation Discord channel, containing the text prompt and parameters for each image.
SAC~\cite{sac} is a dataset of images generated from Stable Diffusion and GLIDE~\cite{glide}, along with user ratings from an aesthetic survey.
However, SAC only contains limited user choices compared to our dataset.

\noindent\textbf{Learning from human feedback.}
Human feedback has long been used in a wide range of deep learning tasks.
Christiano \etal ~\cite{firstrlhf} and Arakawa \etal ~\cite{Arakawa2018DQNTAMERHR} incorporate human feedback into RL training, which is proven to accelerate the model convergence.
Krishna \etal ~\cite{Krishna2022SociallySA} propose ``socially situated AI'', which significantly improves image recognition performance via interacting with human users on Instagram.
InstructGPT~\cite{instructgpt} fine-tunes GPT via a reward function trained on human feedback, establishing the foundation for the success of ChatGPT.
~\cite{Hao2022OptimizingPF} and ~\cite{Lee2023AligningTM} use similar methodology to improve text-to-image models, which are highly related to our work.
In ~\cite{Hao2022OptimizingPF}, this is achieved by augmenting the text prompt.
~\cite{Lee2023AligningTM} is a concurrent work that focuses more on the exact alignment between text and image, while our work shows that the potential of human feedback is far beyond the exact alignment when the feedback takes into account the aesthetic preference of humans.

\begin{figure}
    \centering
    \includegraphics[width=.7\linewidth]{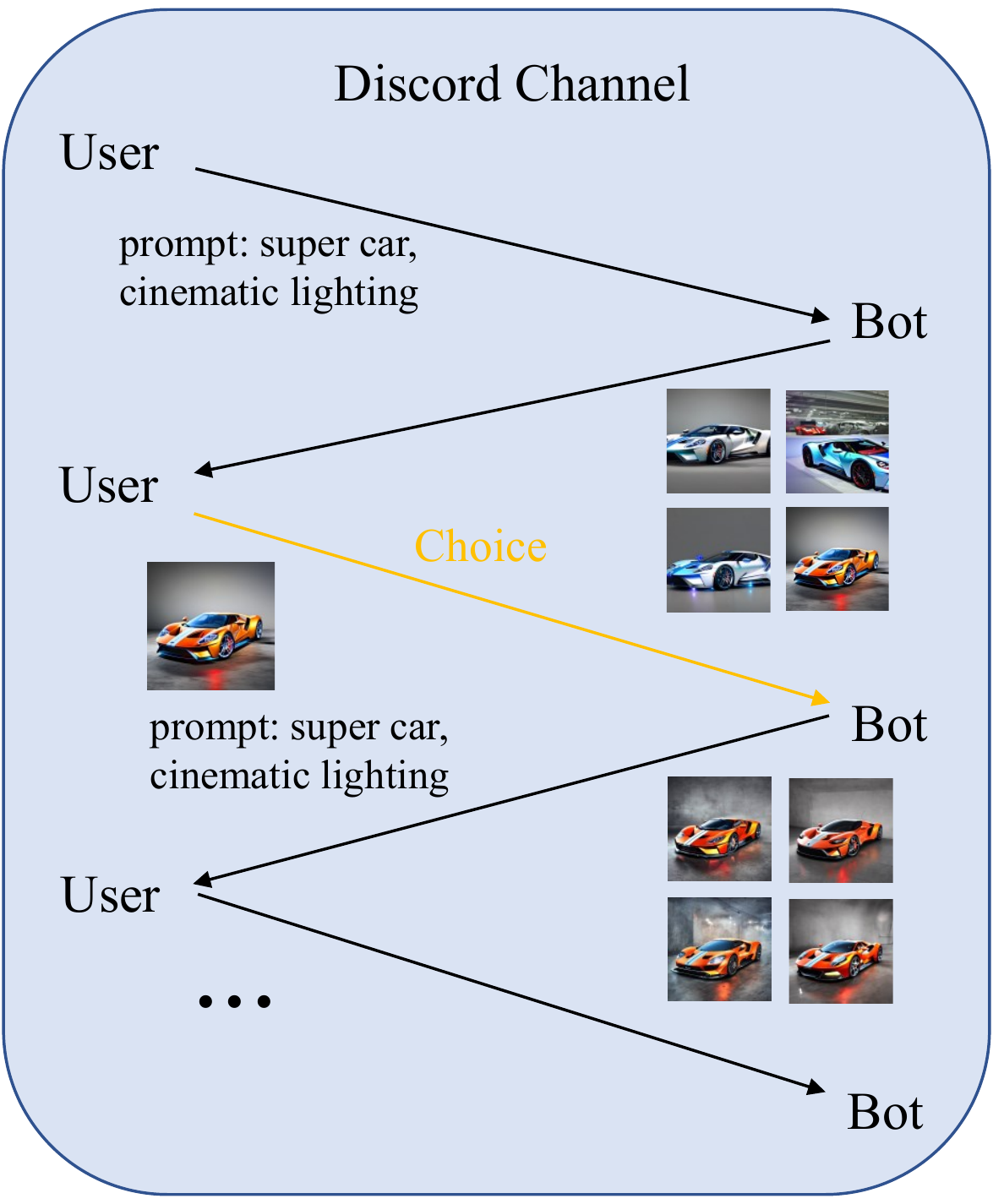}
    \vspace{0.2cm}
    \caption{Interactions in the Discord channel. The human choice is highlighted in orange.}
    \label{fig:data_collect}
\end{figure}

\section{Human Preference Dataset}
\label{sec:dataset}
In order to get a better understanding of the human preferences on the images generated from prompts, and to improve text-to-image generation quality, we start by collecting a dataset of human choices.

\noindent\textbf{Data collection.} 
We utilize the ``dreambot'' channel on the Stable Foundation Discord server to gather human choice data. The chat history of these channels is obtained using the \href{https://github.com/Tyrrrz/DiscordChatExporter}{DiscordChatExporter}~\cite{discord} tool, which downloads the full chat history of a Discord channel and stores it in JSON format. 
Among the chat messages, a discernible pattern of interaction is observed, as depicted in Fig.~\ref{fig:data_collect}, which reveals human preferences.
In this pattern, a user initiates a session by sending a text prompt to the bot, which generates several images in response. 
Then, the user selects a preferred image and sends it back to the bot, along with the original text prompt.
The bot will return several refined images.
This interaction follows a pre-defined grammar, which allows us to extract human choice and related images using simple pattern-matching techniques.

\noindent\textbf{Data format and statistics.} 
Finally, we obtain a total of 98,807 images generated from 25,205 prompts. 
Each prompt corresponds to several images, among which one image is chosen by the user as the preferred one, while others are non-preferred negatives.
Each prompt corresponds with a varying number of images.
23,722 prompts have four images, 953 prompts have three images and 530 prompts have two images. 
The number of images for each prompt depends on the user's specifications in the generation request. Notably, the dataset exhibits a high level of diversity, with images generated across a broad range of themes. 
The dataset consists of choices made by 2,659 different users, and each user contributes at most 267 choices.
Examples of the collected dataset can be found in Fig.~\ref{fig:example}.
For further details on the dataset, we refer the readers to Fig.~\ref{fig:more_example} in the appendix.

\noindent\textbf{Privacy and NSFW contents.} We observe that a small portion of images is generated with image condition (the condition image may be either generated or uploaded by the user).
Since user-uploaded images may contain sensitive information or privacy, we do not include them in our dataset.
For the images with potential NSFW content, we use the channel bot's NSFW detector to filter them out.

In this work, we utilize this dataset to study the existing metrics' correlation with human preferences, which will be introduced in Sec.~\ref{sec:metrics}.
The dataset also serves as the training data for our human preference classifier, which is to be introduced in Sec.~\ref{sec:score}.

\section{Existing Metrics}
\label{sec:metrics}

In this section, we show that the current mainstream evaluation metrics are not well correlated with human preferences on our dataset.

\begin{table}[]
\centering
\begin{tabular}{l|cc}
\toprule
             & IS       & FID                    \\
\midrule
Preferred     & $16.27\pm 0.56$ & 38.2 \\ 
Non-preferred & $16.23\pm 0.53$ & 37.7 \\ 
\bottomrule
\end{tabular}
\vspace{0.3cm}
\caption{The IS and FID of both preferred images and non-preferred images in our collected dataset. The FID is computed by comparing against a subset of images from the LAION-5B dataset that corresponds to the text inputs.}
\label{tab:IS_FID}
\end{table}

\subsection{Metrics by Inception Net}

Inception Score (IS)~\cite{inception_score} and Fréchet inception distance (FID)~\cite{fid} are two popular metrics used to evaluate the quality of generated images. 
Both of them perceive an image through an Inception Net~\cite{inceptionnet} trained on ImageNet~\cite{imagenet}.
In this section, we investigate their correlation with human choices.

\noindent\textbf{Inception Score (IS)} 
measures the quality of generated images by computing the expected KL-divergence between the marginal class distribution over all generated images and the conditional distribution for a particular generated image, using the class probability predicted by the Inception Net. 
This metric is expected to capture both the fidelity and diversity of generated images. 
To determine the correlation between IS and human preferences, we compute IS for both the set of preferred and non-preferred images in our dataset. 
For each setting, we divide 20,000 images into 10 splits and reported the mean and standard deviation of IS computed on them. 
Our results, as shown in Tab.~\ref{tab:IS_FID}, indicate no significant difference between the preferred and non-preferred images.

\noindent\textbf{Fréchet Inception Distance (FID)} 
measures the similarity between the embedding feature of generated and real images. This is achieved by fitting the embedding features into a multivariate Gaussian distribution and computing their Fréchet distance. 
To define the target distribution, FID requires a set of real images. 
However, in the case of images generated from user-provided prompts, such as in our dataset, the target distribution is defined by users' intention, which can only be inferred from text prompts. 
To address this, we randomly sample 10,000 text prompts from our dataset, and for each prompt, we query the LAION~\cite{laion} dataset via the official \href{https://knn.laion.ai/knn-service}{api} to find the closest image, which is taken as a ``pseudo ground truth'' for that prompt. 
This provides a set of real images aligned with the users' intentions. 
We randomly sample 10,000 images from both the preferred and non-preferred split of the collected dataset to compute FID with the real images. 
Our results, as shown in Tab.~\ref{tab:IS_FID}, reveal no significant difference between the preferred and non-preferred images in terms of FID. 
This suggests that FID may not be a reliable metric for evaluating human preference.

\noindent\textbf{Discussion.} 
IS and FID may suffer from the following three issues when evaluating human preference.
Firstly, generated images often contain shape artifacts, as shown in Fig.~\ref{fig:seed}. 
However, classification-based CNNs tend to be biased towards image texture rather than shape~\cite{Geirhos2018ImageNettrainedCA}, making them be likely to ignore shape artifacts in generated images. 
Additionally, the domain gap can pose a problem.
While the evaluation model is trained on real images from ImageNet~\cite{imagenet}, the generated images in our dataset exhibit a wide range of styles and themes, from oil painting portraits to digital art of cyborgs. 
As a result, the ImageNet-trained model may not have meaningful representations for these diverse images~\cite{inceptionscore}. 
Furthermore, these metrics are limited by their single-modal nature, which means that they cannot infer user intentions by accessing prompts, unless the target images are known or provided as we do.

\begin{figure*}
    \centering
    \hspace{0.7cm}
    \includegraphics[width=0.85\linewidth]{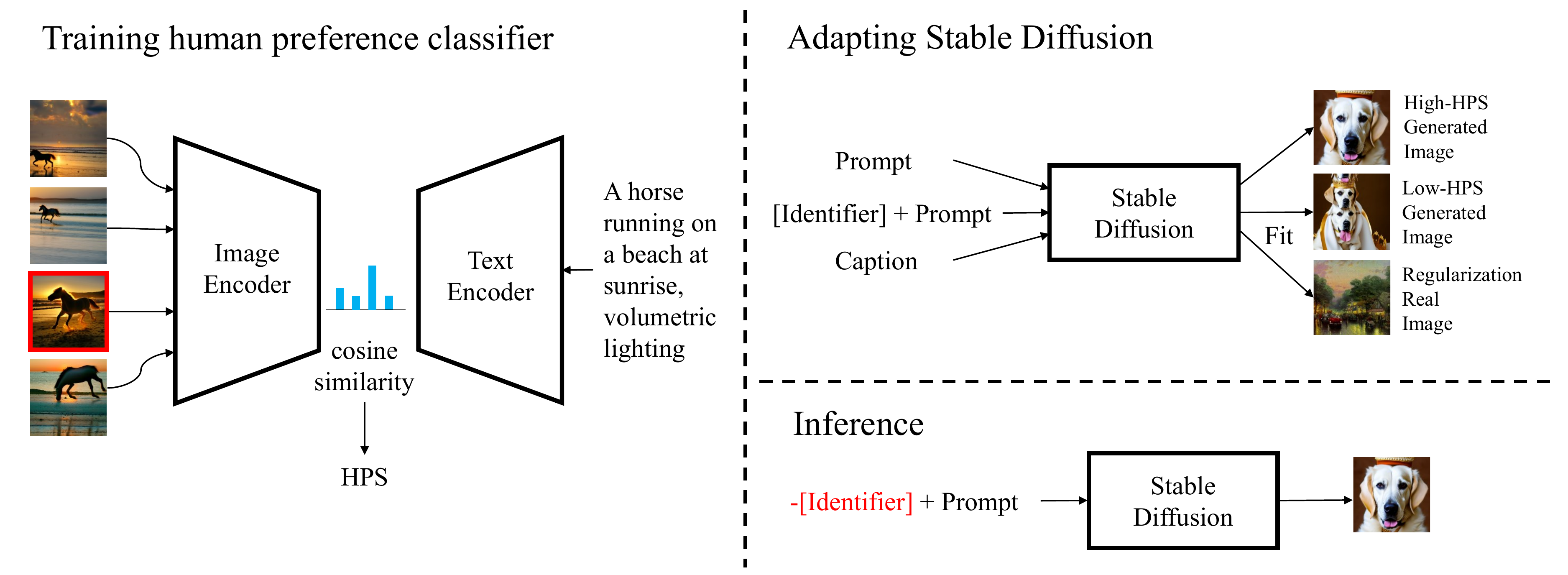}
\caption{Left: training human preference classifier to derive HPS. Right: adapting Stable Diffusion to generate preferable images. During training, the Stable Diffusion is tuned to associate the concept of non-prefer with the prompt prefix [Identifier]. During inference, [Identifier] is used as the negative prompt in classifier-free guidance.}
\label{fig:overview}
\end{figure*}

\subsection{Metrics by CLIP}
\begin{table}[]
\centering
\begin{tabular}{l|c|c}
\toprule
Model                 & Text & Preference Acc. (\%) \\ 
\midrule
Random guess          & \xmark & 26.1  \\ 
Human                 & \cmark & 42.0  \\ 
\midrule
CLIP ViT-L/14         & \cmark & 32.9  \\  
CLIP RN50x64          & \cmark & 33.1  \\ 
Aesthetic Classifier  & \xmark & 31.4 \\  
\midrule
HP classifier (ours)  & \cmark & 43.5 \\   
\bottomrule
\end{tabular}
\vspace{0.3cm}
\caption{Preference Acc. refers to the human choice prediction accuracy on 5,000 user choices. The Aesthetic Classifier makes prediction without seeing the text prompt. }
\label{tab:clip}
\end{table}
Thanks to the large and diverse set of training data, CLIP is better at encoding images from various domains compared to ImageNet-trained models. Moreover, it can capture users' intentions by encoding text prompts, making it a plausible choice for evaluating the alignment between a prompt and a generated image~\cite{glide, dalle2, instructpix2pix, stable_diffusion}.
\href{https://laion.ai/blog/laion-aesthetics/}{Aesthetic Score Predictor} ~\cite{aesthetic_classifier} is another CLIP-based tool for image quality evaluation, which has been utilized to filter the training data for Stable Diffusion~\cite{stable_diffusion}.
In this section, we evaluate the capability of these tools in predicting human choices, which is done by counting the accuracy of the human choice prediction task conducted on a split of 5,000 samples from our dataset.

\noindent\textbf{CLIP score} is derived as the cosine similarity between the prompt embedding and the image embedding computed by CLIP. 
We evaluated the performance of ViT-L/14 and RN50x64 models, which are the largest open-source CLIP models for transformer and CNN architecture. 
Our results, presented in Tab.~\ref{tab:clip}, demonstrate that both CLIP models exhibit superior performance over random guessing. 
However, we will show in Sec.~\ref{sec:reliability} that the CLIP score does not correlate well with human choices. 
Nevertheless, we will also show that it can be further fine-tuned on our dataset to better align with human preferences.

\noindent\textbf{Aesthetic score } is based on a pre-trained ViT-L/14 CLIP image encoder, which is adapted to the task of aesthetic score prediction by adding a MLP layer on top of the CLIP image encoder.
The MLP is trained on several aesthetic datasets, including both real images and generated images (e.g., AVA~\cite{ava}, SAC~\cite{sac}) to predict aesthetic scores ranging from 1 to 10.
Unlike CLIP, the aesthetic classifier does not condition on the prompt, so the image with the highest predicted score is taken as the model choice.
As shown in Tab.~\ref{tab:clip}, the aesthetic classifier also exhibits better-than-chance accuracy in predicting user choice, indicating the importance of the aesthetic aspect of an image in human decision-making.

\section{Human Preference Score}
\label{sec:score}
We first train a human preference classifier to predict the human choice based on the prompt, and then derive HPS based on the trained classifier. 

\noindent\textbf{Human preference classifier} 
We fine-tune the ViT-L/14 version of CLIP on our dataset to better align with human preferences.
Each sample in the training set contains one prompt along with $n \in \{2,3,4\}$ images, among which only one image is preferred by the user.
The model is trained to maximize the similarity between the embedding of the text prompt computed by the CLIP text encoder and the embedding of the preferred image computed by the CLIP visual encoder, while minimizing the similarity for non-preferred images.
By fine-tuning on human choices of generated images, the model is encouraged to better align with human preferences.

\noindent\textbf{Human preference score (HPS)} is derived from the human preference classifier. 
We define HPS as:
$$ \mathrm{HPS}(\mathrm{img}, \mathrm{txt}) = 100 \cos(\mathit{enc}_v(\mathrm{img}), \mathit{enc}_t(\mathrm{txt})),$$
where $\mathit{enc}_v$ and $\mathit{enc}_t$ are the visual encoder and the text encoder of the human preference classifier.
We multiply the cosine similarity by a factor of 100 for better visualization.

\section{Better Aligning Stable Diffusion with Human Preferences}
HPS can be used to guide diffusion-based generative models to better align with human users.
We argue that the misalignment between generated images and human preferences is a problem of missing ``awareness'' rather than model capacity. 
To address this issue, we propose to adapt the generative model by explicitly distinguishing preferred images from non-preferred ones. 
Our solution is straightforward and intuitive.
We construct another dataset consisting of prompts and their newly generated images, which we categorize as either preferred or non-preferred using our previously trained human preference classifier. 
For the non-preferred images, we modify their corresponding prompts by prepending a special prefix. 
By adapting Stable Diffusion on this dataset via LoRA~\cite{lora}, we enhance the model's ability to learn the concept of non-preferred images, which can subsequently be avoided during inference.

\noindent\textbf{Constructing training data.} 
We construct the training data from the ``\texttt{large\_first\_1m}'' split of DiffusionDB~\cite{diffusiondb}, and a subset of the pre-train dataset of Stable Diffusion (LAION-5B) for regularization.
DiffusionDB~\cite{diffusiondb} is a large-scale dataset of generated images along with their text prompts.
For images from DiffusionDB, we first compute HPS for each image-prompt pair.
After that, we group the images by their prompts, and for each prompt $T$, we add the image $I^*$ with the highest HPS into the training data if it passes the following criteria:
$$p > \frac{\alpha}{n},$$
where $n$ is the number of images with the same prompt, and $\alpha$ is a hyper-parameter that controls the selectivity. 
$p$ is given by:
$$ p = \frac{\exp(HPS(I^*, T))}{\sum_{I\in B}\exp(HPS(I, T))},$$
where $B$ is the set of images with the same prompt.
Similarly, we construct the non-preferred subset by the same criteria, but using negative HPS.
Finally, we get a mixed dataset of generated images and real images, where the non-preferred generated images are identified by their prompt prefix.

\noindent\textbf{Adapting Stable Diffusion.} We adopt LoRA~\cite{lora} to adapt Stable Diffusion to the training data, in which the parameters of the original model are kept frozen, and the \{key, query, value, out\} projection matrices are augmented with a low-rank residual.
LoRA does not add new parameters to the model, since the learned projection matrices can be merged into the base model once trained.
During training, we use the prompt as the caption for generated images.
For non-preferred images, we prepend a special identifier before each of their captions (we choose ``Weird image.'' as the special identifier in our case).
During inference, the special identifier is used as the negative prompt for classifier-free guidance~\cite{classifierfree} to avoid generating non-preferred images.

\begin{figure}
    \centering
    \includegraphics[width=0.75\linewidth]{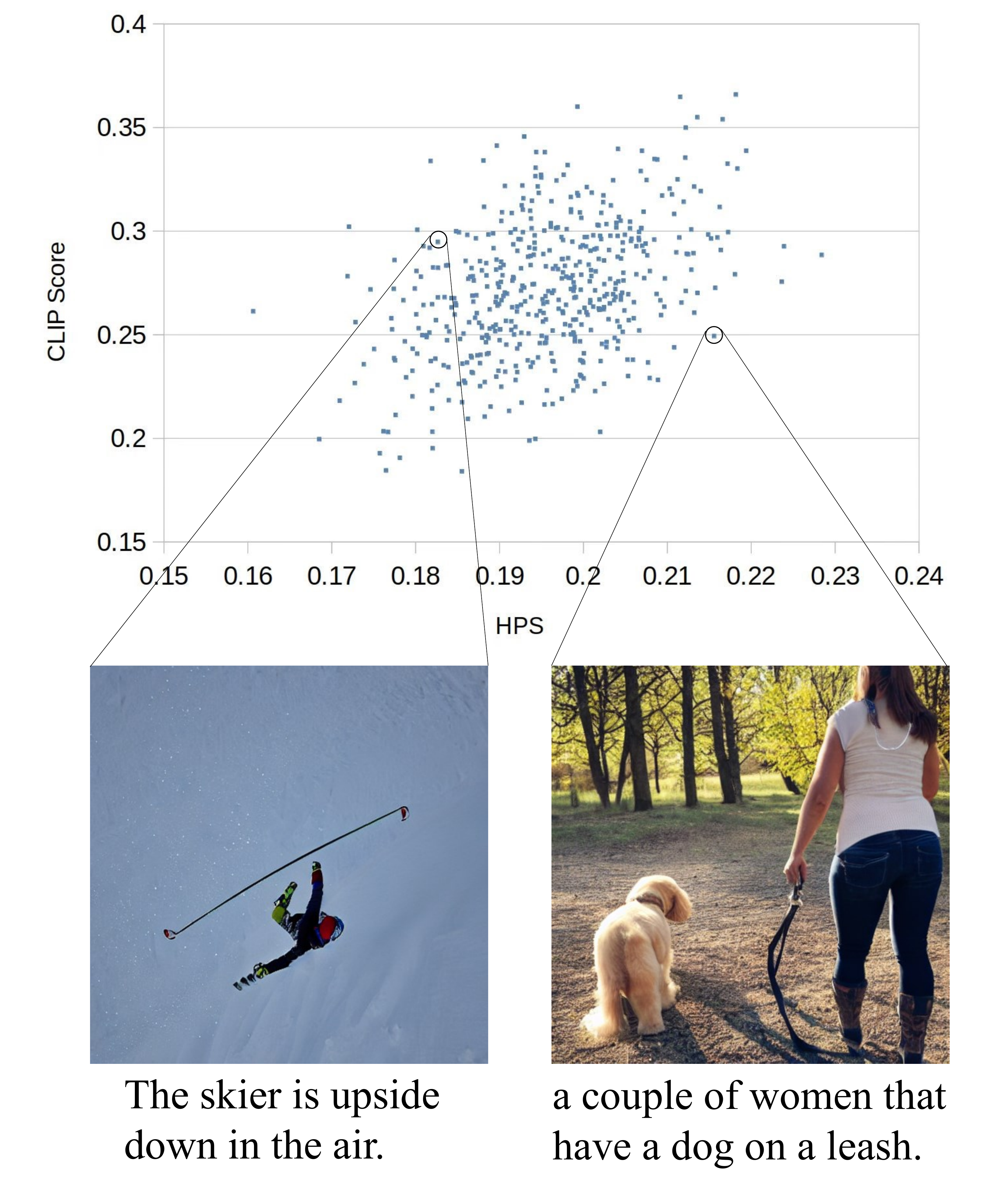}
    \caption{Correlation between HPS and CLIP score. While the CLIP score emphasizes more on the direct matching between the image content and the text prompt, HPS emphasizes more on the aesthetic quality of images.}
    \label{fig:vis}
\end{figure}

\section{Experiments}

In this section, we firstly validate the reliability of HPS in Sec.~\ref{sec:reliability}, and then in Sec.~\ref{sec:application}, we introduce our experiments of adapting Stable Diffusion.

\begin{figure*}
    \centering
    \includegraphics[width=0.9\linewidth]{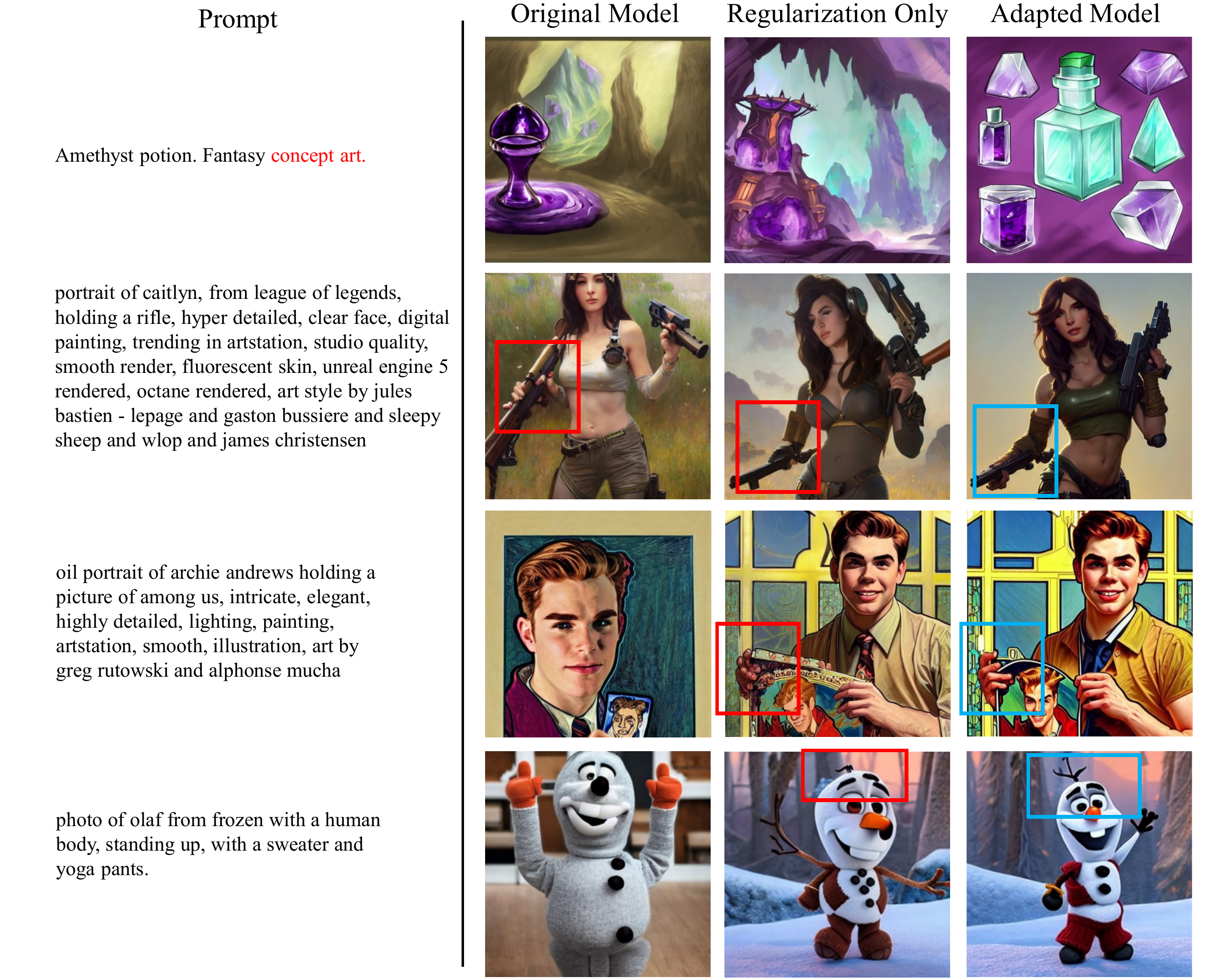}
\caption{Comparison of images generated by the original model, the regularization-only model, and our adapted model. ``Regularization Only'' refers to a head-to-head setting against the ``Adapted Model'', where Stable Diffusion is adapted without HPS-labeled images. Images in the same row are generated with the same prompt and random seed. The prompts are sampled from DiffusionDB. The adapted model can better capture the user intention from the prompt, and generate more preferable images with fewer artifacts.}
\label{fig:quality}
\end{figure*}

\subsection{HPS}
\label{sec:reliability}
\noindent\textbf{Implementation details of human preference classifier. }
We use 20,205 samples from our dataset during training, which contains 20,205 prompts and 79,167 images.
We use the ViT-L/14 version of CLIP in our experiments.
We fine-tune the last 10 layers of the CLIP image encoder and the last 6 layers of the text encoder.
The model is trained by the AdamW optimizer~\cite{adam} with a learning rate of $1.7\times10^{-5}$ for 1 epoch.
The batch size is 5.
The learning rate decays with a cosine learning rate schedule.
Weight decay is set as $3.1\times10^{-3}$.
Instead of using the original data augmentation of random resized crop, we directly resize the longest edge of the image to 224, and then pad zeros to make the shorter edge increase to 224.
We empirically find that fixing the aspect ratio of the image is beneficial.
The hyper-parameters are tuned via Bayesian optimization.

\begin{table}[]
\centering
\begin{tabular}{l|c}
\toprule
                & Agreement (\%) \\
\midrule
Human vs. Human & $63.5 \pm 4.3$  \\
CLIP vs. Human  & $56.8 \pm 1.7$  \\
\midrule
HPS vs. Human   & $61.5 \pm 1.1$  \\
\bottomrule
\end{tabular}
\vspace{0.4cm}
\caption{Agreement on comparing images generated by  Stable Diffusion and DALL·E.}
\label{tab:agreement}
\end{table}

\noindent\textbf{Alignment with human. }
As shown in Tab.~\ref{tab:clip}, the trained model significantly outperforms CLIP in the human choice prediction task.
Due to the strong diversity of human preferences, the accuracy is even higher than our human participants.

\noindent\textbf{Generalization. } We evaluate HPS' generalization capability towards other generative models by user studies.
In this experiment, we let the human preference classifier and several human participants evaluate 398 pairs of images.
In each pair, the images are generated by DALL·E~\cite{dalle2} and Stable Diffusion~\cite{stable_diffusion} with the same text prompt.
The prompts are randomly sampled from DiffusionDB~\cite{diffusiondb}, which is a large database of images and prompts sourced from the Stable Foundation Discord channel.
We filter out the NSFW prompts by the indicator provided in DiffusionDB~\cite{diffusiondb}.

In Tab.~\ref{tab:agreement}, we evaluate the agreement between the predictions from humans, CLIP, and HPS.
The agreement is computed by averaging the similarity of the prediction of each participant.
HPS is better aligned with human preference compared to CLIP score, and its agreement with humans is close to the agreement between humans.
It shows that HPS can generalize toward images generated by other models.
We refer the readers to the supplementary material for a full list of images and choices made in this user study.

\noindent\textbf{Correlation with CLIP score. }
In Fig.~\ref{fig:vis}, we visualize the correlation between HPS and CLIP score.
The text prompts are randomly sampled from the COCO Captions~\cite{cococaptions} dataset, and the images are generated by Stable Diffusion~\cite{stable_diffusion}.
We can see that HPS has a positive correlation with CLIP score, but emphasizes more on the aesthetic quality of an image.
However, HPS put less importance on the direct matching between image contents and text prompts, which can be interpreted as a visual analogy of ``alignment tax'' introduced in ~\cite{instructgpt}.

\subsection{Better Aligning Stable Diffusion with Human Preferences}
\label{sec:application}
\noindent\textbf{Implementation details. }
We use the Stable Diffusion~\cite{stable_diffusion} v1.4 for all our experiments.
$\alpha$ is set to 2.0 for both preferred images and non-preferred images when constructing the training set.
The constructed training set contains 37,572 preferred generated images and 21,108 non-preferred generated images.
The regularization images are from a 625k subset of LAION-5B filtered by the aesthetic score predictor with a threshold of 6.5.
200,231 regularization images participate in training.
We only fine-tune the UNet of Stable Diffusion, while keeping the VAE and the text encoder frozen during training.
The rank is set to 32 in LoRA~\cite{lora}. 
The LoRA weights are trained for 10k iterations with the AdamW~\cite{adam} optimizer with a learning rate of $1\times10^{-5}$ and a weight decay of $1\times10^{-2}$, which is kept constant during training.
We use a batch size of 40 in our experiments.
For inference, we run the diffusion process by 50 steps for each image with PNDM~\cite{Liu2022PseudoNM} noise scheduler.
We use the default guidance scale of 7.5 for classifier-free guidance~\cite{classifierfree}.

\begin{figure}
    \centering
    \includegraphics[width=0.8\linewidth]{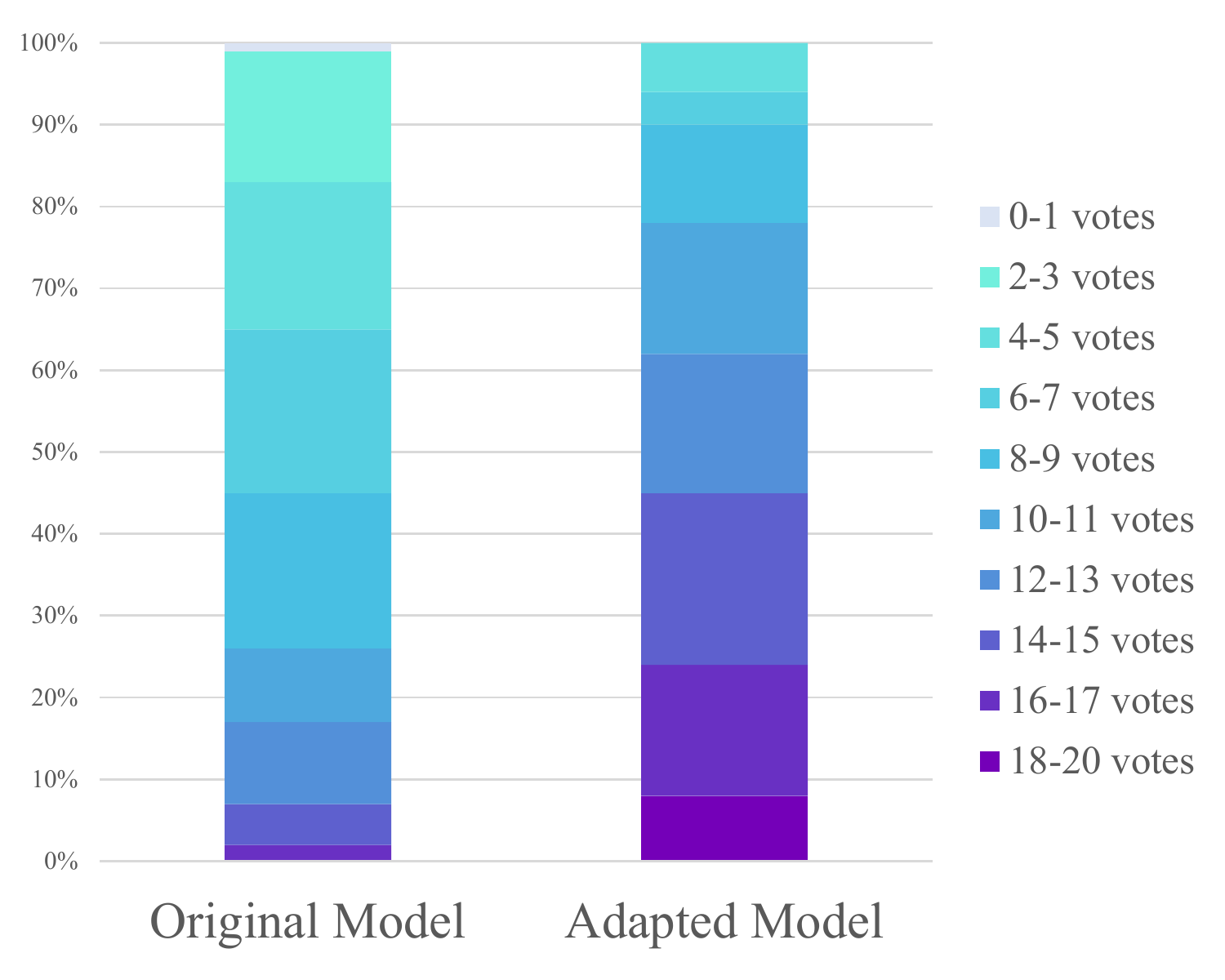}
\caption{Human evaluation results on images generated from 100 randomly sampled prompts. The color represents the number of positive votes received from 20 participants.}
\label{fig:user_study}
\end{figure}

\noindent\textbf{Human evaluation. }
We compare our trained model with the original Stable Diffusion by conducting user studies.
In this study, we randomly sample 100 user-provided prompts from DiffusionDB~\cite{diffusiondb}.
For each prompt, we generate an image from both models with the same random seed for fair comparison, resulting in 100 pairs of generated images for the user study.
We ask 20 participants to read the prompt, and then choose between the image generated by our trained model and the original Stable Diffusion based on their preference.
In Fig.~\ref{fig:example}, we visualize our result by showing the percentages of images with different numbers of positive votes.
The adapted model significantly outperforms the original model.
74\% of the images generated by the adapted model has more than 10 votes, while the number is 22\% for the original model.
A screenshot of the user-study interface is presented in Fig.~\ref{fig:screenshot} in the appendix.

\noindent\textbf{Qualitative Evaluation.}
In Fig~\ref{fig:quality}, we show some typical cases of improvement.
We compare the original model, the regularization-only model, and the adapted model.
The adapted model is trained with both real regularization images and generated images with HPS preference labels.
The regularization-only model is a head-to-head comparison with the adapted model, which is trained by removing the generated images from the training set and is trained exclusively on regularization images for the same number of steps.
The results show that the adapted model can better capture the user intention from the prompt, as shown in the first row.
The last three rows show that training with generated images mitigates the problem of unnatural limbs.
We refer the readers to Fig.~\ref{fig:quality} and Fig.~\ref{fig:artifacts} in the appendix for more examples.

\begin{table}[]
\centering
\resizebox{1.0\linewidth}{!}{
\begin{tabular}{l|cccc}
\toprule
&  FID $\downarrow$ & Aesthetic Score~\cite{aesthetic_classifier} $\uparrow$  & CLIP Score~\cite{clip} $\uparrow$ & HPS $\uparrow$  \\
\midrule
SD 1.4     & 19.72 & 5.90 & 0.2816 & 0.1898 \\
Adapted model & 19.35 & 6.06 & 0.2831 & 0.1916 \\
\bottomrule
\end{tabular}
}
\vspace{0.3cm}
\caption{Comparison between the original SD v1.4 and the adapted model.}
\label{tab:comparison}
\end{table}
\noindent \textbf{Quantitative Evaluation.}
In Tab.~\ref{tab:comparison}, we compare the adapted model with the baseline on FID, Aesthetic Score, CLIP Score and HPS.
The FID~\cite{fid} is computed on 10k images from the LAION~\cite{laion} dataset. 
CLIP Score~\cite{clip} and HPS are computed on prompts from DiffusionDB~\cite{diffusiondb}.

\section{Limitations}
\label{sec:limitations}
There are several limitations about the dataset.
The collected dataset contains generated prompts and images of public figures. We choose to mark them out instead of removing them to keep the diversity of the dataset.
Despite the diversity of the dataset, we are also aware that it only represents the preference of a small portion of people in the world, and it may be biased towards a certain group of people that are active in the Stable Foundation Discord channel.
Another potential bias about this dataset is that a large portion of text prompts are written by experienced Stable Diffusion users.
These prompts are very likely to be tweaked to activate the potential of Stable Diffusion and deviate from normal language habits.

\section{Conclusion}
In this work, we study human preferences on a large-scale dataset of generated images.
We find that the previous evaluation metrics for generative models are not well aligned with human preferences, but the CLIP model can be fine-tuned into a human preference classifier to better align with human choices.
Then, we show a simple yet effective method to adapt the generative model to generate more preferable images with the guidance of human preference score.
We hope our work can inspire the community to explore new possibilities of human-aligned AI research.

\section*{Acknowledgement}
This project is funded in part by National Key R\&D Program of China Project 2022ZD0161100, by the Centre for Perceptual and Interactive Intelligence (CPII) Ltd under the Innovation and Technology Commission (ITC)’s InnoHK, by General Research Fund of Hong Kong RGC Project 14204021. Hongsheng Li is a PI of CPII under the InnoHK.
This project is also supported by SenseTime Collaborative Research Grant.

{\small
\bibliographystyle{ieee_fullname}
\bibliography{egbib}
}

\newpage

\appendix

\clearpage

\section{Datasheet}
\label{sec:datasheet}

\subsection{\centering \centering Motivation}
\subsection*{Why was the dataset created?} 
\noindent The dataset was created to facilitate future academic Computer Vision research about human aesthetic preference.
\subsection*{Who created this dataset (\eg which team, research group) and on behalf of which entity (\eg company, institution, organization)?}
\noindent The dataset was created by researchers at MMLab, The Chinese University of Hong Kong.

\subsection{\centering Composition}
\subsection*{What do the instances that comprise the dataset represent (\eg
documents, photos, people, countries)? Are there multiple types of instances? (\eg movies, users, ratings; people, interactions between them; nodes, edges)}
\noindent
The instances are prompts and generated images, along with human preference choices among the images generated by the same prompt.

\subsection*{Are relationships between instances made explicit in the data (\eg social network links, user/movie ratings, etc.)?}
\noindent
Yes, instances generated by the same user are identified by the same user id, which is anonymized for privacy.

\subsection*{How many instances are there? (of each type, if appropriate)?}
\noindent 
There are 25,205 instances in the dataset.

\subsection*{What data does each instance consist of? ``Raw'' data (\eg unprocessed text or images) or Features/attributes? Is there a label/target associated with instances? If the instances related to people, are sub-populations identified (\eg by age, gender, etc.) and what is their distribution?}
\noindent 
Each instance consists of $n\in {2,3,4}$ image, one prompt and one human choice. 

\subsection*{Is any information missing from individual instances? If so, please
provide a description, explaining why this information is missing (\eg
because it was unavailable). This does not include intentionally removed
information, but might include, \eg redacted text.}
\noindent 
Yes, we omit the specific parameters for generating the images, such as diffusion steps and guidance scale.
They are omitted because we are more interested in the users' preference about the generated images, rather than how they are created.
Also, since the same batch of images (among which users make comparisons) are always generated with the same set of parameters except the random seed, they are irrelevant variables when studying human preferences.

\subsection*{Is everything included or does the data rely on external resources?}
\noindent The dataset is self-contained.

\subsection*{Are there recommended data splits and evaluation measures? (\eg training, development, testing; accuracy or AUC)}
\noindent 
In our experiments, we use a training set of 20,205 instances and validation set of 5,000 images, which will be made public.
We recommend using accuracy (\%) with one decimal place.

\subsection*{Are there any errors, sources of noise, or redundancies in the dataset?}
\noindent Yes. The users are not prompted to selected images fitting their preference, so there should be noise in the collected data.

\subsection*{Is the dataset self-contained, or does it link to or otherwise rely on external resources (\eg websites, tweets, other datasets)?}
\noindent The dataset is self-contained.

\subsection*{Does the dataset contain data that might be considered confidential (\eg data that is protected by legal privilege or by doctorpatient confidentiality, data that includes the content of individuals non-public communications)?}
\noindent 
No, the dataset is collected from the Stable Foundation Discord server, which is publicly available for any user with an account.

\subsection*{Does the dataset contain data that, if viewed directly, might be offensive, insulting, threatening, or might otherwise cause anxiety? If so, please describe why.}
\noindent
We collect images and their prompts from the Stable Foundation discord server. 
Even though the discord server has rules against users sharing any NSFW (not suitable for work, such as sexual and violent content) and illegal images, our dataset still contains some NSFW images and prompts that were not removed by the server moderators.

\subsection*{Does the dataset relate to people?}
\noindent Yes, the prompts are written by users and the choices are made by users.

\subsection*{Does the dataset identify any subpopulations (\eg by age, gender)?}

\noindent No.

\subsection*{Is it possible to identify individuals (\ie one or more natural persons), either directly or indirectly (\ie in combination with other data) from the dataset?}
\noindent No.

\subsection*{Does the dataset contain data that might be considered sensitive in any way (\eg data that reveals racial or ethnic origins, sexual orientations, religious beliefs, political opinions or union memberships, or locations; financial or health data; biometric or genetic data; forms of government identification, such as social security numbers; criminal history)?}

The dataset may contain sensitive data, because the prompts written by users may contain sensitive information, such as public figures and religious beliefs.

\subsection*{What experiments were initially run on this dataset? Have a summary of those results.}
\noindent 
It has been used to validate the correlation between human preference and several popular image quality evaluation metrics, and serve as the training data for a human preference classifier.
The results show that the tested metrics do not correlate well with human preference, and the correlation of the ViT-L/14 version of CLIP can be improved via fine-tuning on the dataset.


\subsection{\centering Data Collection Process}

\subsection*{How was the data associated with each instance acquired?}

\noindent
The prompts, images and human choices are directly observable from the Stable Foundation Discord server.

\subsection*{What mechanisms or procedures were used to collect the data (\eg hardware apparatus or sensor, manual human curating, software program, software API)?}
\noindent Automatic scraping procedures were used to collect the data.

\subsection*{If the dataset is a sample from a larger set, what was the sampling strategy (\eg deterministic, probabilistic with specific sampling probabilities)?}

\noindent  The dataset is not a sample of a larger set.

\subsection*{Who was involved in the data collection process (\eg students, crowd-workers, contractors) and how were they compensated (\eg
how much were crowdworkers paid)?}

\noindent The authors of this paper were solely involved in the data collection process.

\subsection*{Over what time-frame was the data collected?}
\noindent 
The dataset covers the chat history of dreambot channels between Dec. $2^{nd}$ 2022 and Jan. $18^{th}$ 2023.

\subsection*{Were any ethical review processes conducted (\eg by an institutional review board)?}
\noindent No official processes were conducted, due to the public nature of the data on Discord channel.

\subsection*{Does the dataset relate to people?}
\noindent No.

\subsection*{Did you collect the data from the individuals in question directly, or obtain it via third parties or other sources (\eg websites)?}
\noindent 
The data was obtained from public messages in the Discord server.





\subsection*{Has an analysis of the potential impact of the dataset and its use on data subjects (\eg a data protection impact analysis)been conducted?}

\noindent No analysis has been conducted.


\subsection{\centering Data Preprocessing}
\subsection*{What preprocessing/cleaning was done? (\eg discretization or bucketing, tokenization, part-of-speech tagging, SIFT feature extraction, removal of instancess, processing of missing values)?}
\noindent No preprocessing is done on the images and prompts. 




\subsection{\centering Uses}
\subsection*{Has the dataset been used for any tasks already? If so, please provide a description.}
\noindent As described in the paper, this dataset has been used for analysis about several image quality evaluation metrics and training the proposed human preference classifier.

\subsection*{Is there a repository that links to any or all papers or systems that use the dataset?}
\noindent
No.

\subsection*{What (other) tasks could the dataset be used for?}
\noindent 
It can be used for tasks related to human preference on generated images.

\subsection*{Is there anything about the composition of the dataset or the way it was collected and preprocessed/cleaned/labeled that might impact future uses?}
\noindent Yes. 
As discussed in Sec.~\ref{sec:limitations}, the dataset is biased towards the preference of the certain group of people that are active in the Stable Foundation Discord server.


\subsection*{Are there tasks for which the dataset should not be used?}
\noindent No.


\subsection{\centering Data Distribution}
\subsection*{Will the dataset be distributed to third parties outside of the entity (\eg company, institution, organization) on behalf of which
the dataset was created? If so, please provide a description}
\noindent Yes. Researchers at academic institutions will be able to request access to the dataset. 

\subsection*{How will the dataset be distributed? (\eg tarball on website, API, GitHub; does the data have a DOI and is it archived redundantly?)}
\noindent We will provide download links for researchers on a GitHub repository. 

\subsection*{When will the dataset be distributed?}
\noindent Before April 15, 2023.

\subsection*{Will the dataset be distributed under a copyright or other intellectual property (IP) license, and/or under applicable terms of use (ToU)?}
\noindent We will provide a terms of use agreement with the dataset. The dataset as a whole will be distributed under a non-commercial license. 

\subsection*{Have any third parties imposed IP-based or other restrictions on
the data associated with the instances? If so, please describe these restrictions, and provide a link or other access point to, or otherwise reproduce, any relevant licensing terms, as well as any fees associated
with these restrictions.}
\noindent No.

\subsection*{Do any export controls or other regulatory restrictions apply to
the dataset or to individual instances? If so, please describe these
restrictions, and provide a link or other access point to, or otherwise
reproduce, any supporting documentation.}
\noindent Unknown.


\subsection{\centering Dataset Maintenance}
\subsection*{Who is supporting/hosting/maintaining the dataset?}
\noindent The authors of this paper are maintainers of this dataset.

\subsection*{How can the owner/curator/manager of the dataset be contacted
(\eg email address)?}
\noindent By email: wuxiaoshi@link.cuhk.edu.hk .

\subsection*{Is there an erratum?}
\noindent At this time, we are not aware of errors in our dataset. However, we will create an erratum as errors are identified.

\subsection*{Will the dataset be updated? If so, how often and by whom? 
How will updates be communicated? (\eg mailing list, GitHub)}
\noindent 
The dataset will be updated by the authors on an at-will basis (but no more than once a month).

\subsection*{If the dataset relates to people, are there applicable limits on the
retention of the data associated with the instances (\eg were individuals in question told that their data would be retained for a
fixed period of time and then deleted)? If so, please describe these
limits and explain how they will be enforced.}
\noindent No such limits are established.

\subsection*{Will older versions of the dataset continue to be supported/hosted/maintained?}
\noindent N/A

\subsection*{If others want to extend/augment/build on this dataset, is there a mechanism for them to do so? If so, is there a process for tracking/assessing the quality of those contributions. What is the process for communicating/distributing these contributions to users?}
\noindent There will not be a mechanism to build on top of the dataset.


\begin{figure*}
    \centering
    \includegraphics[width=1.0\linewidth]{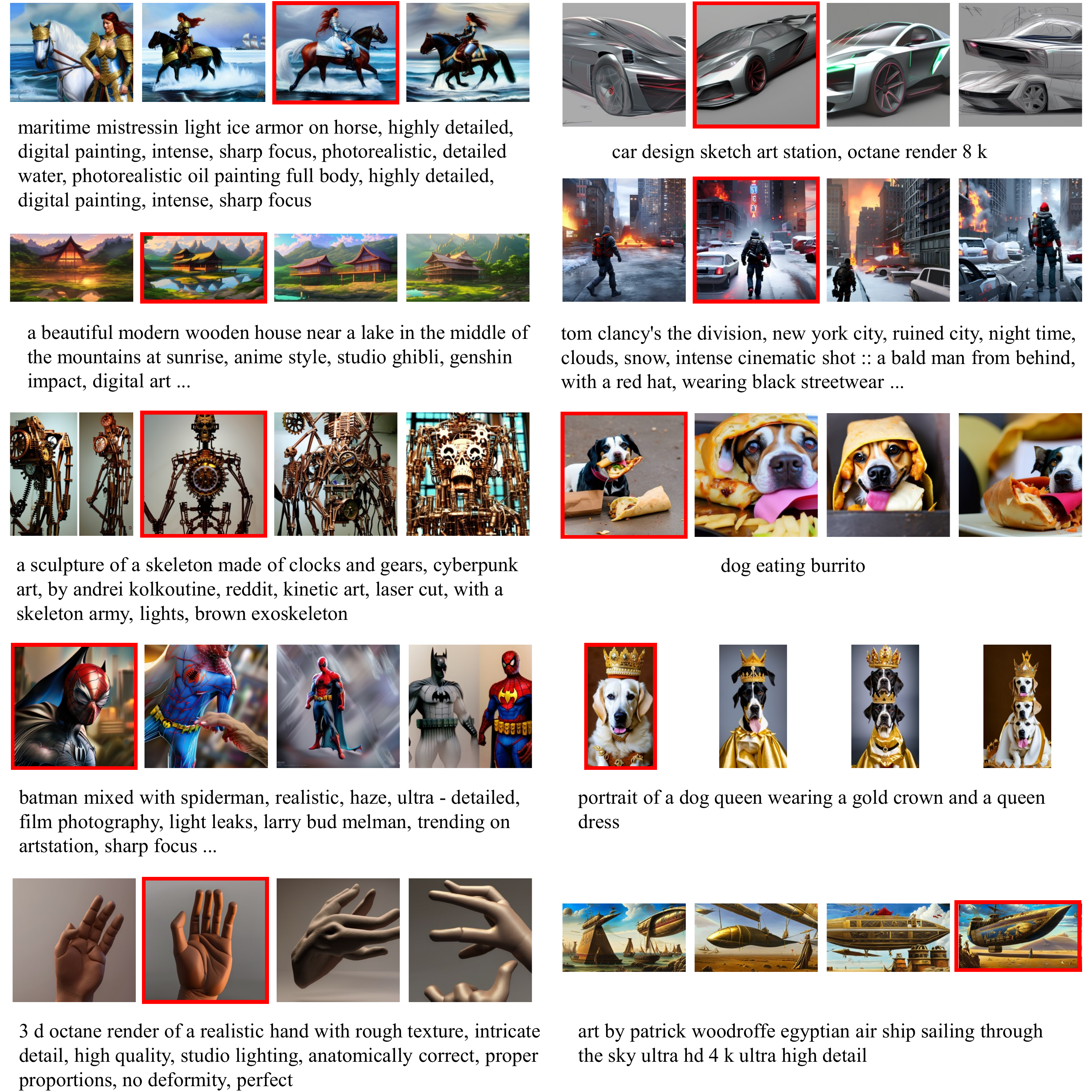}
\vspace{0.2cm}
\caption{More examples of the collected data. The images are generated by Stable Diffusion with the prompts shown below each group of images. The preferred images are highlighted with red borders. }
\label{fig:more_example}
\end{figure*}

\begin{figure*}
    \centering
    \includegraphics[width=0.9\linewidth]{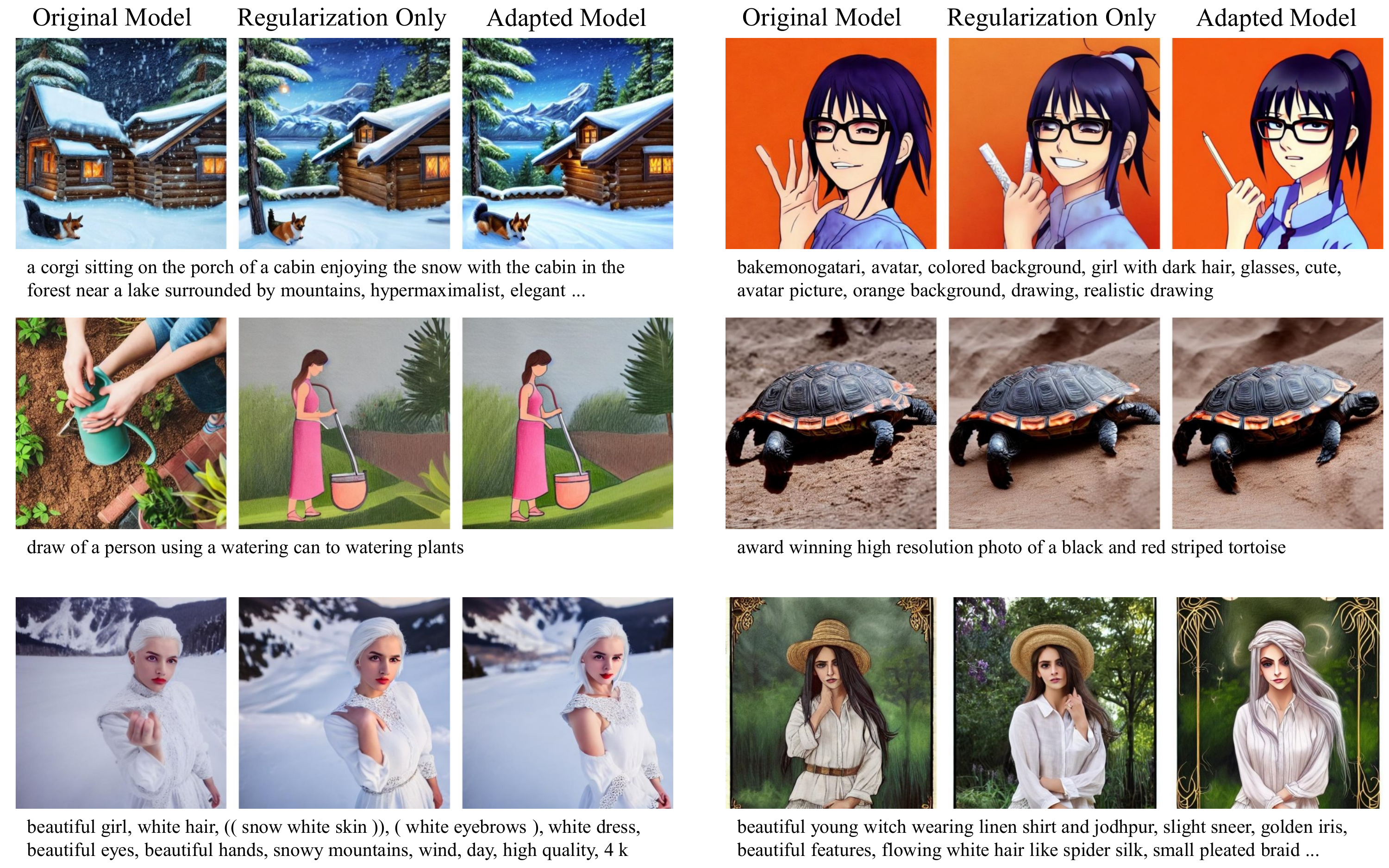}
\vspace{0.2cm}
\caption{The adapted model generates images with less artifacts. Images in the same group are generated with the same prompt and random seed.}
\label{fig:artifacts}
\vspace{0.5cm}
    \includegraphics[width=0.9\linewidth]{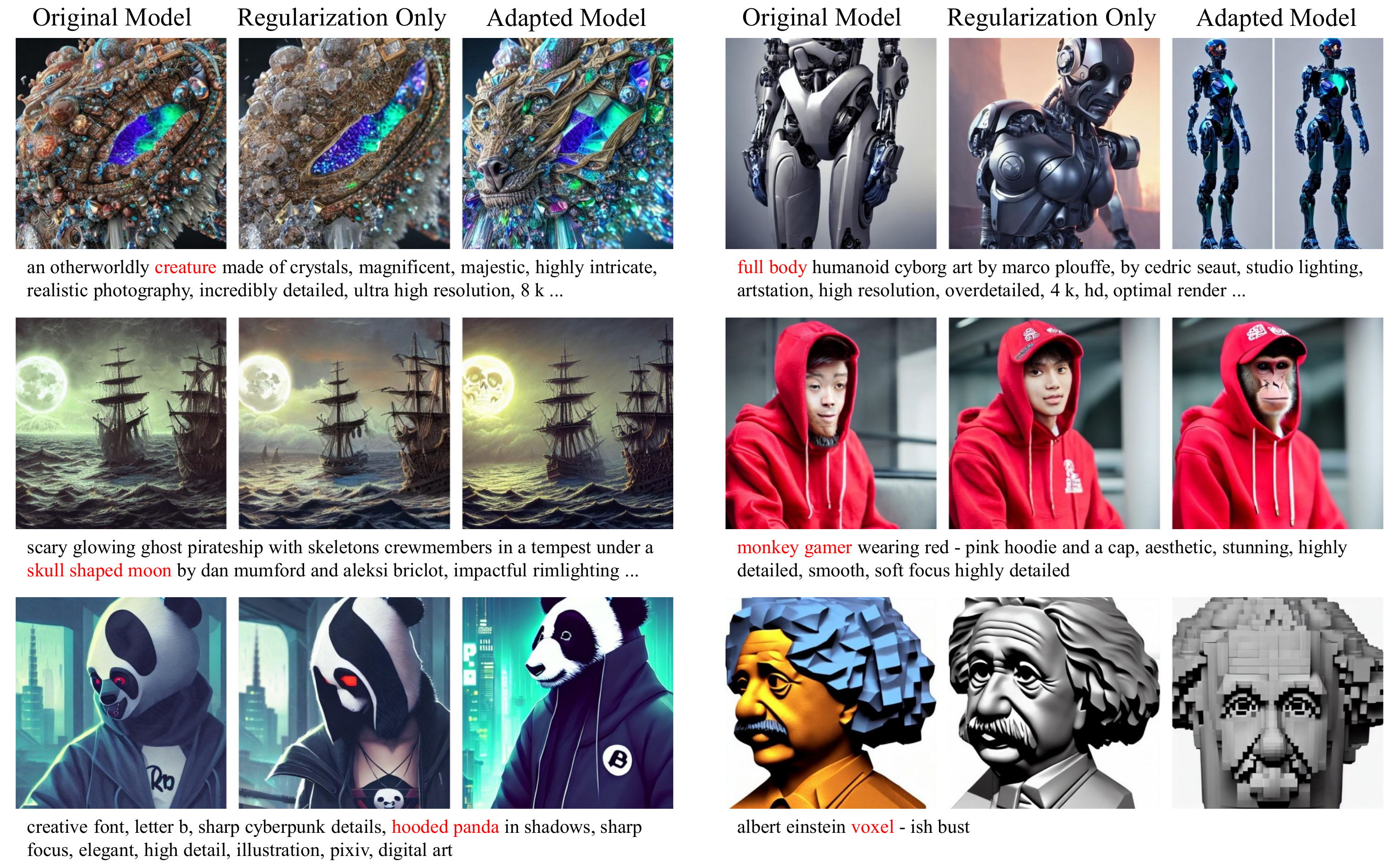}
\vspace{0.2cm}
\caption{The adapted model generates images that better capture user intentions. Images in the same group are generated with the same prompt and random seed.}
\label{fig:intention}
\end{figure*}

\section{More Dataset Examples}
\label{sec:dataset_example}
See Fig.~\ref{fig:more_example} for more examples.

\section{More Visualization}
\label{sec:adapted_model_vis}
See Fig.~\ref{fig:artifacts} and Fig.~\ref{fig:intention} for more visualizations.
We show that the adapted model generates images with less artifacts and are better aware of users' intentions.

\label{sec:screenshot}
\begin{figure*}
    \centering
    \includegraphics[width=0.7\linewidth]{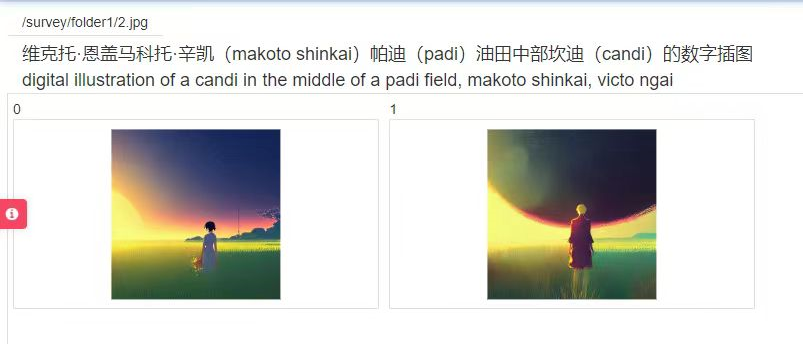}
\vspace{0.2cm}
\caption{Screenshot of the user-study interface.}
\label{fig:screenshot}
\end{figure*}

\end{document}